\documentclass[10pt,twocolumn,letterpaper]{article}

\usepackage{wacv}              

\usepackage{graphicx}
\usepackage{amsmath}
\usepackage{amssymb}
\usepackage{booktabs}

\usepackage[table,xcdraw,dvipsnames]{xcolor}

\usepackage{pifont}
\usepackage{amssymb}

\usepackage{multirow}
\usepackage{diagbox}
\usepackage{makecell}
\usepackage{amssymb}
\usepackage{booktabs} 
\usepackage{setspace}

\usepackage[capitalize]{cleveref}
\crefname{section}{Sec.}{Secs.}
\Crefname{section}{Section}{Sections}
\Crefname{table}{Table}{Tables}
\crefname{table}{Tab.}{Tabs.}

\def\wacvPaperID{1191} 
\def\confName{WACV}
\def\confYear{2025}

\begin{document}


\title{
Retaining and Enhancing Pre-trained Knowledge \\in Vision-Language Models with Prompt Ensembling
} 


\author{
Donggeun Kim\textsuperscript{1,2}\footnotemark[1] \footnotemark[2]\;\;\;\;  Yujin Jo\textsuperscript{1}{\footnotemark[1]}\;\;\;\;  Myungjoo Lee\textsuperscript{1}{\footnotemark[1]}\;\;\;\;  Taesup Kim\textsuperscript{1}\footnotemark[3]\\
\textsuperscript{1}Graduate School of Data Science, Seoul National University\;\;\;\; 
\textsuperscript{2}Nota Inc.\\
}

\maketitle

\renewcommand{\thefootnote}{\fnsymbol{footnote}} 
\footnotetext[1]{Equal contribution.} %
\footnotetext[2]{Work done at Seoul National University.}
\footnotetext[3]{Corresponding author.}
\renewcommand{\thefootnote}{\arabic{footnote}} 

\begin{abstract}
The advancement of vision-language models, particularly the Contrastive Language–Image Pre-training (CLIP) model, has revolutionized the field of machine learning by enabling robust zero-shot learning capabilities. These capabilities allow models to understand and respond to previously unseen data without task-specific training.
However, adapting CLIP to integrate specialized knowledge from various domains while retaining its zero-shot capabilities remains a significant challenge. 
To address this, we introduce a novel prompt ensemble learning approach called Group-wise Prompt Ensemble (GPE). This method aims to enhance CLIP's zero-shot capabilities by incorporating new domain knowledge while improving its adaptability and robustness against data distribution shifts.
Our approach hinges on three main strategies: prompt grouping with masked attention to optimize CLIP's adaptability while safeguarding its zero-shot capabilities; the incorporation of auxiliary prompts for the seamless integration of new domain insights without disrupting the original model's representation; and an ensemble learning strategy that effectively merges original and new knowledge.
Through rigorous experimentation, including more challenging cross-dataset transfer evaluations, our GPE method redefines the benchmarks for the adaptability and efficiency of vision-language models, surpassing existing models across various scenarios.
\end{abstract}

\begin{figure}
    \centering
    \includegraphics[width=1\linewidth]{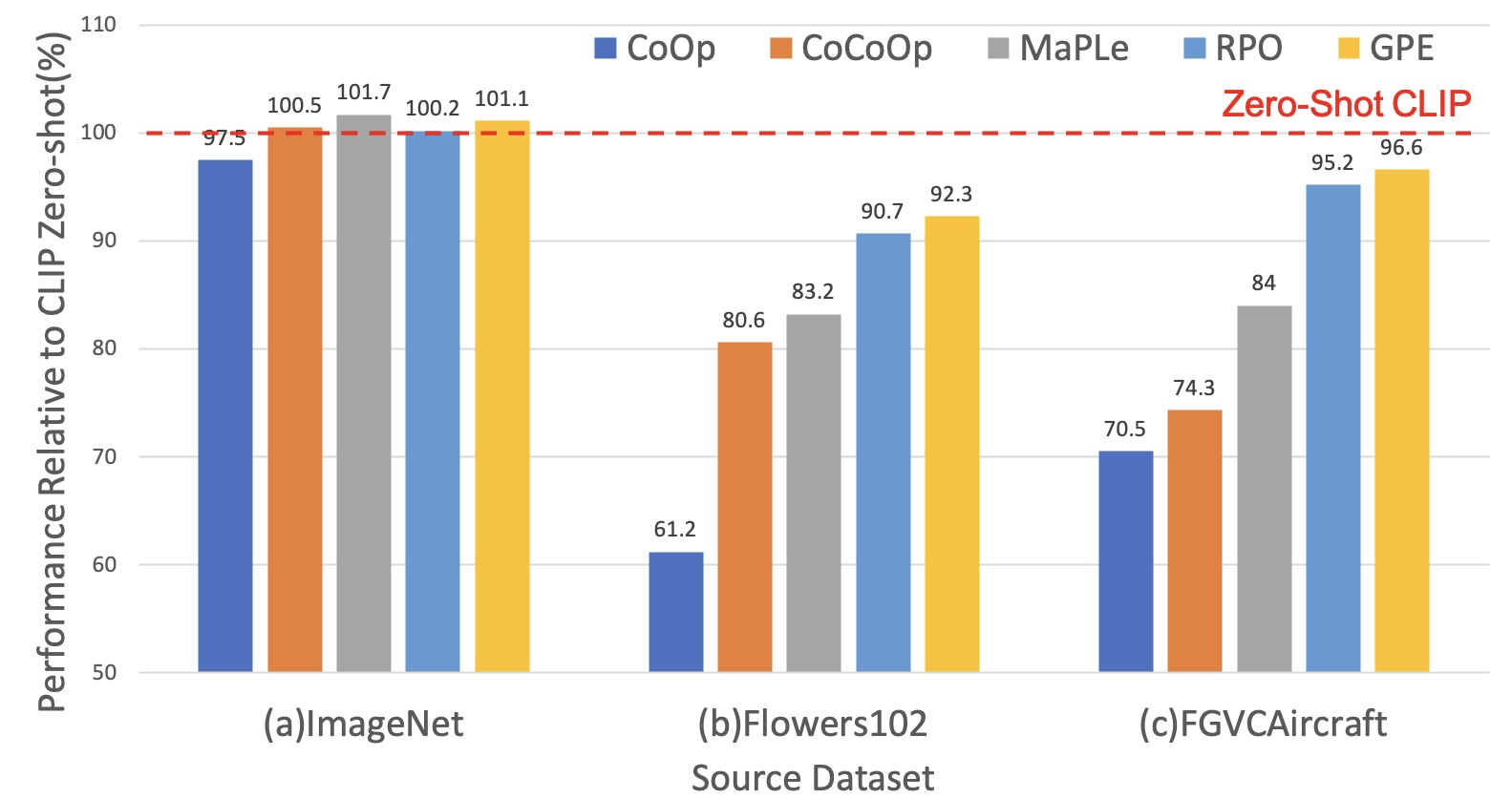}
    \caption{\textbf{Cross-Dataset Evaluation across various source datasets.} This evaluation measures how well models trained on a specific source dataset (e.g., ImageNet, Flowers102, FGVCAircraft) generalize when tested on 10 other target datasets, relative to CLIP’s zero-shot performance. When trained on a general dataset like (a), most models maintain or even exceed CLIP's zero-shot performance. However, when fine-tuned on specialized datasets like (b) and (c) and evaluated on other datasets, baseline models show significant performance drops. In contrast, our model, GPE, demonstrates strong performance even on these niche datasets, highlighting its ability to adapt without losing generalization.}
    
    \label{fig:motivation_xd}
\end{figure}
\vspace{-10pt}

\section{Introduction}
\label{sec:intro}

Vision-language models (VLMs), such as CLIP~~\cite{clip} and ALIGN~~\cite{align}, have exhibited exceptional proficiency in understanding the complex interactions between visual and textual information. These models, powered by transformers~~\cite{transformer} and trained on extensive datasets of paired images and texts through contrastive learning, have demonstrated impressive zero-shot learning capabilities, allowing them to recognize and interpret unseen concepts without the need for additional fine-tuning.

However, adapting these models to specific domains or datasets not encountered during pre-training remains challenging. Fine-tuning vision-language models often leads to a noticeable degradation in their original zero-shot performance~\cite{wortsman2022robust}. In response, prompt-based methods~\cite{coop, cocoop} have emerged as a parameter-efficient strategy to preserve a  vision-language models' pre-trained knowledge by avoiding changes to its weights during adaptation.

Despite this progress, a critical issue we identified is that CLIP struggles to maintain its zero-shot capabilities when fine-tuned on specialized or fine-grained datasets, as shown in Figure \ref{fig:motivation_xd}. While existing prompt learning methods~\cite{coop, cocoop, maple, rpo} perform well on generic datasets like ImageNet~\cite{imagenet}, they suffer a significant drop in performance when fine-tuned on domain-specific datasets about flower categories\cite{flowers} or aircraft model variants\cite{aircraft} and subsequently evaluated on other visual classification datasets. This highlights a major limitation in their adaptation to specialized data while preserving generalization across diverse tasks.

To address this challenge, we propose a novel method, the Group-wise Prompt Ensemble (GPE), designed to preserve the zero-shot capabilities of CLIP while adapting them to specific domains. GPE ensures the model remains effective across diverse datasets, even after fine-tuning on niche domains. The key innovation of GPE lies in diversifying the learning process through prompt grouping, which maintains the model's zero-shot capabilities while incorporating new domain-specific knowledge.
Our approach is based on three main strategies:
\begin{itemize}
    \item We implement prompt grouping with attention masks to enhance domain adaptability while ensuring no disruption to the model's original representation.
    \item We introduce auxiliary prompts to effectively integrate new domain insights and enrich the learning context.
    \item We use an ensemble learning strategy that merges original and new knowledge by promoting diversity among prompts. This ensures each prompt contributes unique and complementary information to the final prediction.
\end{itemize}
These strategies significantly improve CLIP’s generalization and adaptability across challenging scenarios. In evaluation, the GPE method excels in base-to-novel class generalization, outperforming zero-shot CLIP in novel class accuracy and demonstrating superior performance on diverse datasets. Additionally, in our proposed extended cross-dataset evaluations, our model showcases remarkable resilience against out-of-distribution datasets, achieving near zero-shot performance levels even after fine-tuning on specific datasets. These findings underline the GPE's effectiveness in preserving and utilizing distinct knowledge bases, setting a new benchmark for applying vision-language models in real-world scenarios.

\begin{figure*}[t!]
\centering
\includegraphics[width=0.9\textwidth]{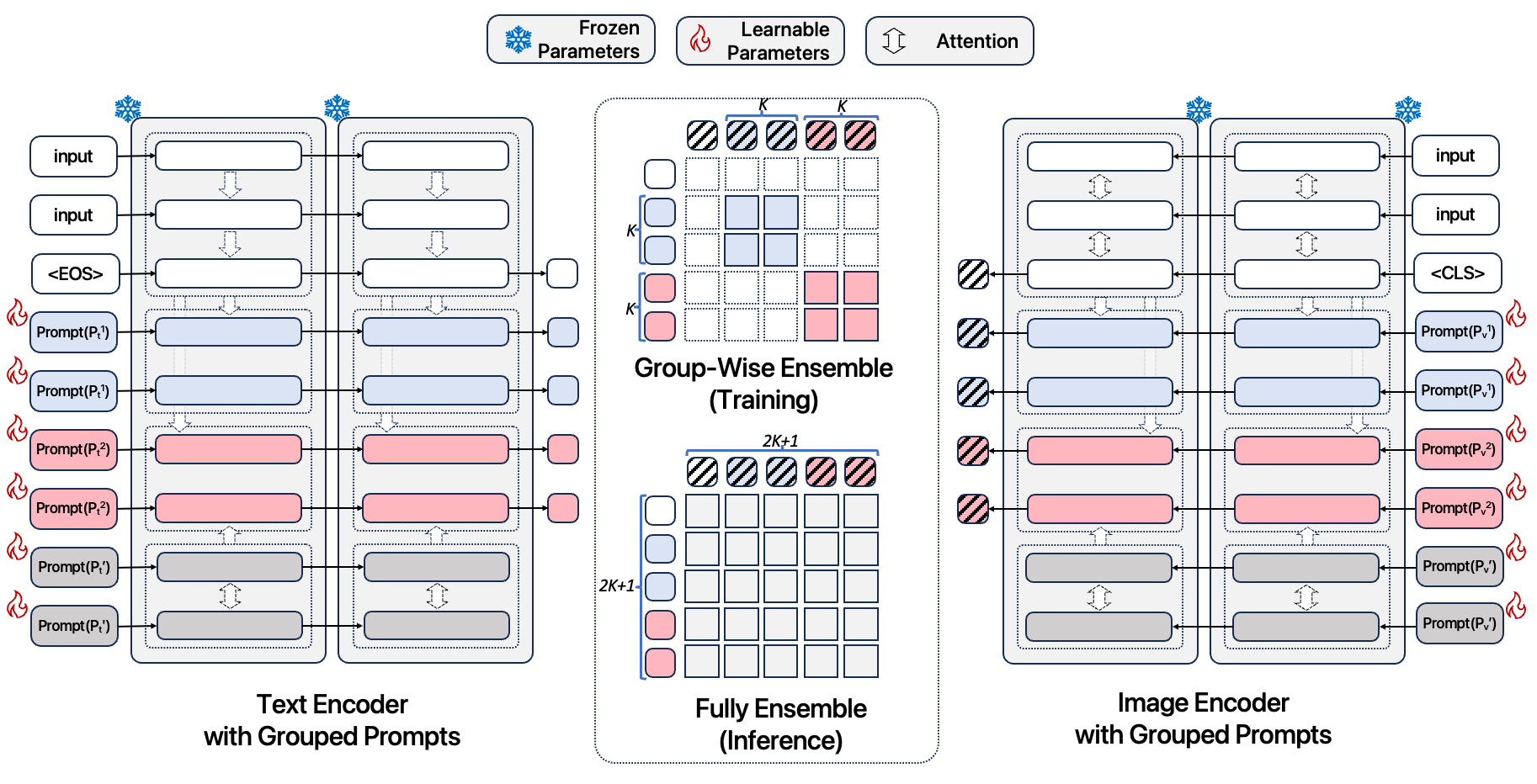}
\caption{\textbf{Overview of our framework.} The framework consists of a Text Encoder with Grouped Prompts($P_{t}$) and an Image Encoder with Grouped Prompts($P_{v}$). The first group of the main prompts($P_{t}^{1}$, $P_{v}^{1}$)  is shown in blue, the second group($P_{t}^{2}$, $P_{v}^{2}$) in red, and the auxiliary prompts($P_{t}^{\prime}$, $P_{v}^{\prime}$) in gray. During training, we utilize a Group-wise Ensemble approach, while for inference, we employ a Full Ensemble strategy. The Full Ensemble Inference process effectively integrates diverse insights from all prompt groups and the special token (white) from CLIP to enhance predictive performance.}
\label{fig:overview}
\end{figure*}
\vspace{-10pt}

\section{Related Work}
\subsection{Vision-Language Models}
Vision-language models (VLMs)~\cite{clip, align, zhai2022lit, yuan2021florence, yao2021filip} demonstrate exceptional performance, with CLIP~\cite{clip} and ALIGN~\cite{align} being prominent examples. These models have gained recognition for their remarkable ability to comprehend images and texts in a zero-shot manner. Their success owes not only to the development of transformers~\cite{transformer}  but also to the training on large amounts of image-text data. CLIP is trained on 400 million image-text pairs, while ALIGN utilizes 1.8 billion pairs, enabling them to obtain versatile image representations through contrastive learning. This method identifies matching pairs as positive and mismatches as negative, equipping VLMs with a solid understanding of open-vocabulary concepts for a various applications. 

However, fine-tuning these models on specific datasets often undermines their zero-shot efficacy. Recent efforts~\cite{wortsman2022robust} focus on tailoring pre-trained VLMs to specific tasks, aiming to retain their inherent strengths while optimizing for particular challenges.

\subsection{Prompt Learning}
Prompt learning originated from the field of natural language processing (NLP), stemming from advancements in models like GPT~\cite{gpt,gpt3} and BERT~\cite{bert}. It emerged as an alternative to traditional fine-tuning methods, by integrating additional tokens into pre-trained language models. This strategy, often in the form of prompts or instructions, aims the efficient adaptation for downstream tasks without updating pre-trained weights.

Recently, there has been a surge of interest in extending prompt-based methods to vision-language models, with notable efforts including studies on~\cite{coop,cocoop,maple,proda,promptsrc,prograd}. However, due to the fewer parameters used in training prompt-based models than full fine-tuning, conveying sufficient knowledge to the model poses a challenge, often requiring knowledge to be crammed into prompts. Our model is designed to address this issue by diversifying prompts and employing an ensemble strategy. This dual approach allows the model to acquire a wide range of information, improving its overall performance.

\subsection{Ensemble Learning}

Averaging predictions across multiple models can significantly boost a predictive system's generalization ability. This technique, known as ensemble methods, has been widely adopted in the field of machine learning~\cite{dietterich2000ensemble,deep_ensemble}. In the realm of vision-language models, the approach of prompt ensembling usually utilizes multiple prompts to create an ensemble of classifiers within the logit space, where each classifier is defined by its unique prompt. For example, the textual prompt augmentation has shown promise in improving zero-shot performance~\cite{zeroshot_prompt_ensemble,tpt} and enhancing the model's generalization capability~\cite{promptsrc,apollo}. In prompt tuning,~\cite{rpo} demonstrate an ensemble effect by averaging the logits of independent pair-wise prompts.

However, a significant challenge in prompt ensemble lies in information redundancy among prompts. The recent approaches~\cite{barlow,vicreg} in self-supervised learning tackle a similar issue by decorrelating output units. This is achieved by minimizing the off-diagonal elements of the cross-correlation matrix, ensuring each output provides unique information. We adopt this regularization strategy to increase diversity in prompt predictions and reduce redundancy.

\section{Methods}
\subsection{Preliminaries} CLIP is built on two encoders, one for image and the other for text. Since we use a transformer~\cite{transformer} based architecture in this work, CLIP encodes an image and a corresponding text description as described below:
\begin{equation*}
\begin{aligned}
& \mathbf{x}^{(0)}=\left[x^{(0)} ; E_x^{(0)} ;P_v\right], 
& \mathbf{y}^{(0)}=\left[y^{(0)} ; E_y^{(0)} ;P_t\right]
\end{aligned}
\end{equation*}

where $x^{(0)} \in \mathbb{R}^{d_v}$ and $y^{(0)} \in \mathbb{R}^{d_t}$ denote special token embeddings, \texttt{[CLS]} for the image encoder and \texttt{[EOS]} for the text encoder respectively, which serves as feature aggregators in each encoder. $E_x^{(0)} \in \mathbb{R}^{N_x \times d_v}$ and $E_y^{(0)} \in \mathbb{R}^{N_y \times d_t}$ denote the image and text embeddings, where $N_x$ and $N_y$ denote the length of feature tokens (not counting the special token) in image and text, and $d_v, d_t$ are the dimensions of image patch and word embeddings. Text embeddings $E_y^{(0)}$ are encoded using a prompt template such as \texttt{‘A photo of a [CLASS\_NAME]'}. For prompt tuning, continuous learnable prompts $P_{v}$ for vision and $P_{t}$ for language are appended to each input token sets.

\subsection{Prompt Grouping with Auxiliary Prompts}\label{sec:group-wise ensemble}


\begin{figure*}[t!]
\centering
\includegraphics[width=0.8\textwidth]{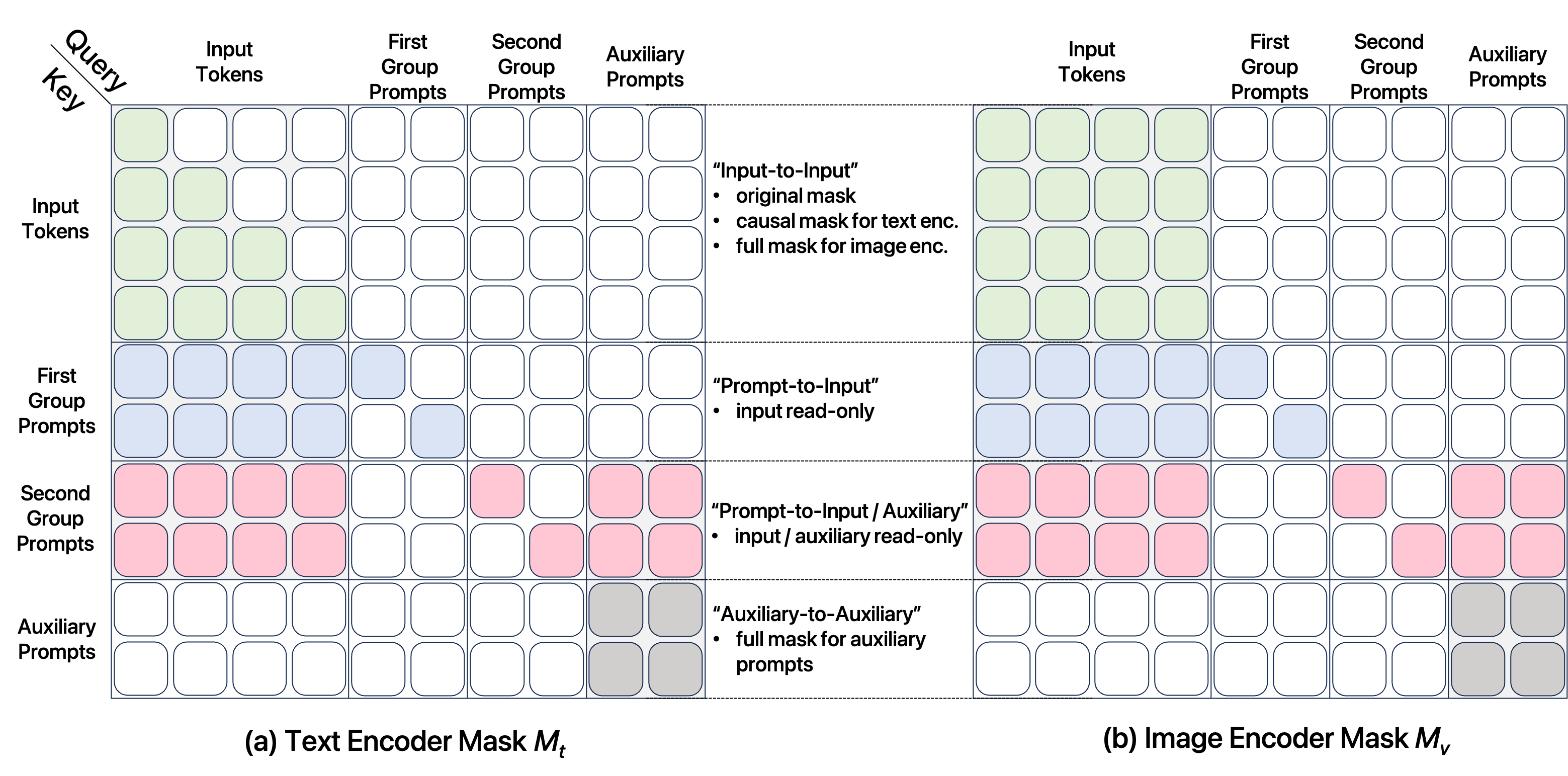}
\caption{\textbf{Attention masks of GPE.} In transformer models, attention masks determine which input parts can interact by allowing or blocking connections between them. The colored boxes indicate areas where attention occurs, while the white boxes indicate masked regions. The first group prompts are restricted to reading the input only, without modifying it. In the second group prompts, masking allows attention to both the input and auxiliary. Each prompt token can perform self-attention by attending to itself. The auxiliary prompts attend only to themselves, with all other features remaining masked.} 
\label{fig:mask}
\end{figure*}
\vspace{-5pt}


\noindent{\textbf{Grouped Prompts and Auxiliary Prompts}} We introduce Group-wise Prompt Ensemble (GPE), designed to preserve CLIP's inherent zero-shot capabilities while integrating domain-specific knowledge. As illustrated in Figure \ref{fig:overview}, we divide the prompts into three sets: $P_{v}=\left[P_{v}^{1};P_{v}^{2};P_{v}^{\prime}\right]$ and $P_{t}=\left[P_{t}^{1};P_{t}^{2};P_{t}^{\prime}\right]$, with a total length of $K_\text{total}$. The prompts $P^{1}$ and $P^{2}$, each with an equal length of $K$, are initialized with \texttt{[CLS]} or \texttt{[EOS]} as in previous work~\cite{rpo}. Each prompt token serves as the special token, optimized for downstream task prediction. $P^{1}$ is referred to as the 'first group', and $P^{2}$ is denoted as the 'second group'. Together, they form the 'main prompt'. 

 Unlike main prompts, auxiliary prompts $P^{\prime}$ do not participate in predictions but are used to aid in training the second group. By setting auxiliary prompts only to attend with second group, we ensure that each group of main prompts evolves in a distinct manner during training. This approach combines complementary predictions from a model ensemble perspective, resulting in improved overall performance. $P^{\prime}$ are randomly initialized, with a length of $K^{\prime}$. Therefore, the total prompt length is defined as $K_\text{total}=K+K+K^{\prime}$. 

\noindent{\textbf{Masked Attention}} First, similar to RPO\cite{rpo}, we use read-only prompts that prevent the original features from being altered by the learnable prompt embeddings, thus preserving the model’s pre-trained knowledge and minimizing internal representation shifts. Second, we introduce masked attention within the framework of prompt grouping. The first group is restricted to reading the input only without modifying it. For the second group, masking allows attention to both the input tokens and auxiliary prompts. The auxiliary prompts can attend to itself, while all other features remain masked except for itself. Our masked attention strategy not only minimizes internal shifts but also allows multiple prompt groups to be trained independently. This leads to both greater prompt diversity, and more robust ensemble effect, improving overall performance.

Image attention mask $M_v \in \mathbb{R}^{N_v \times N_v}$ and textual attention mask $M_t \in \mathbb{R}^{N_t \times N_t}$ are illustrated in Figure \ref{fig:mask}. Here, $N_v=1+N_x+K_\text{total}$ and $N_t=1+N_y+K_\text{total}$. Then, the attention mask can be defined as follows, where $M^{i, j}$ refers the $i$ th row, $j$ th column element:
\begin{equation*}
\begin{aligned}
M_v^{i, j}= \begin{cases}
0, & \text{if }  i \leq1+N_x+2K \text{ and } j\leq1+N_x \\
         & \text{or }  1+N_x < i=j \leq 1+N_x+2K \\
         & \text{or }  i> 1+N_x+K  \text{ and }  j>1+N_x+2K\\
-\infty, & \text{otherwise}
\end{cases}
\end{aligned}
\vspace{-8pt}
\end{equation*}
\begin{equation*}
\begin{aligned}
M_t^{i, j}= \begin{cases}
0, & \text{if } j\leq i \leq1+N_y+2K \text{ and } j\leq1+N_y \\
         & \text{or }  1+N_y < i=j \leq 1+N_y+2K \\
         & \text{or }  i> 1+N_y+K \text{ and } j>1+N_y+2K  \\
-\infty, & \text{otherwise}
\end{cases}
\end{aligned}
\end{equation*}

The process of masked attention within the transformer encoder can be described as follows:
\begin{equation*}
\begin{aligned}
&\mathrm{x}^{(l+1)}  =\mathcal{V}_{l+1}\left(\mathrm{x}^{(l)}, M_v\right) ,
&\mathrm{y}^{(l+1)}  =\mathcal{T}_{l+1}\left(\mathrm{y}^{(l)}, M_t\right) 
\end{aligned}
\end{equation*}
where $\mathcal{V}_{l+1}$ and $\mathcal{T}_{l+1}$ denote the $(l+1)$-th multi-head self-attention layer of the image encoder and text encoder with each attention mask, while $\mathbf{x}^{(l)} \in \mathbb{R}^{N_v \times d_v}$, and $\mathbf{y}^{(l)} \in \mathbb{R}^{N_t \times d_t}$ are input tokens for each layer. The outputs from the last transformer block $L$, $ \mathbf{x}^{(L)}$ and $\mathbf{y}^{(L)}$, are as follows:
\begin{equation*}
\begin{aligned}
\mathbf{x}^{(L)}=\left[e_0 ; E_x^{(L)} ;\left\{e_i\right\}_{i=1}^{K};\left\{e_i\right\}_{i=K+1}^{2K};e^{\prime}\right],\\
\mathbf{y}^{(L)}=\left[s_0 ; E_y^{(L)} ;\left\{s_j\right\}_{j=1}^{K};\left\{s_j\right\}_{j=K+1}^{2K};s^{\prime}\right]
\end{aligned}
\end{equation*}
where $ e_i$ and $s_j$, each are combined with the special token and the main prompts, while $ e^{\prime}$ and $s^{\prime}$ are auxiliary prompts. The final representations used for training and inference are obtained by projecting $ e_i$ and $s_j$ into a text-image joint embedding space, using projection matrix $\mathbf{P}_v$ and $\mathbf{P}_t$:
\begin{equation}
\begin{aligned}
V&=\mathbf{P}_v \cdot e_i=\left[v_0 ; \left\{v_i\right\}_{i=1}^{K};\left\{v_i\right\}_{i=K+1}^{2K}\right],\\
T&=\mathbf{P}_t \cdot s_j=\left[t_0 ; \left\{t_j\right\}_{j=1}^{K};\left\{t_j\right\}_{j=K+1}^{2K}\right]
\end{aligned}
\label{eq:final_output}
\end{equation}
\subsection{Group-wise Learning with Covariance Loss}\label{sec:group-wise ensemble}
\noindent{\textbf{Group-wise Ensemble}} As shown in Figure \ref{fig:overview}, for $K$ prompts of the same group in each modality, we compute $K\times K$ logits for each group based on cosine similarity given a single image $x$ and class label $y$. We define the prediction probability $p(y=y_k | x) = \text{softmax}\left(\{\operatorname{sim}(x, y_{k'})/ \tau\}_{k'}\right)_k$ for each group of prompts by averaging intra-group logits independently as:
\begin{equation}
\resizebox{0.9\linewidth}{!}{$
\operatorname{sim}(x, y) =
\left\{
\begin{array}{ll}
\frac{1}{K^2} \sum_{i=1}^{K}\sum_{j=1}^{K} \frac{v_i \cdot t_j}{|v_i||t_j|}, & \text{for Group 1}, \\
\frac{1}{K^2} \sum_{i=K+1}^{2K}\sum_{j=K+1}^{2K} \frac{v_i \cdot t_j}{|v_i||t_j|}, & \text{for Group 2}.
\end{array}
\right.
$}
\label{eq:presoftmax}
\end{equation}
where $\tau$ denotes the temperature parameter. Finally, prompts are optimized with the group-wise cross-entropy loss $L_{\text{CE-Group1}}$ and $L_{\text{CE-Group2}}$. 

In our approach, grouping is crucial for boosting the impact of the ensemble. The strength of ensemble lies in the variety it introduces, leading to better performance. By applying loss separately to each group, we ensure that the characteristics of one group significantly differ from those of the other. Each group's distinct and complementary contributions, fostered by grouping strategy through masked attention, enable GPE to exploit a broader range of knowledge and insights.

During training, we apply the pre-softmax method, which averages the raw output values (logits) before applying the softmax function, as shown in Eq~\ref{eq:presoftmax}. For inference, we use the more common post-softmax method, where the softmax probabilities are averaged after calculation, as explained in Eq~\ref{eq:postsoftmax}. We chose the pre-softmax for training because our experiments demonstrated that it provides better performance.

\noindent{\textbf{Covariance Regularization}}
To further enhance the ensemble process, we introduce a covariance regularization. This technique promotes diversification among the feature embeddings during training, ensuring that each prompt contributes unique and complementary information to the final prediction. By doing so, we not only mitigate the risk of information redundancy but also significantly enhance the model's adaptability to new domains, providing a sense of reassurance.

We denote $Z=\left[v_1, \ldots, v_{2K}\right]$ and $Z^{\prime}=\left[t_1, \ldots, t_{2K}\right]$ as the prompts composed of $2K$ vectors of dimension $d$, which are the embeddings generated from each encoder of CLIP. Consequently, we define the covariance matrix of $Z$ as:
\begin{equation*}
    C(Z)=\frac{1}{2K-1}\sum_{i=1}^{2K}\left(v_i-\bar{v}\right)\left(v_i-\bar{v}\right)^T \text{where } \bar{v}=\frac{1}{2K} \sum_{i=1}^{2K} v_i 
\end{equation*}

Inspired by~\cite{barlow}, we define the covariance regularization term $r$ as the sum of the squared off-diagonal coefficients of $C(Z)$, with a factor $1 / d$ that scales the criterion as a function of the dimension:
$r(Z)=\frac{1}{d} \sum_{i \neq j}[C(Z)]_{i, j}^2$
This term encourages the off-diagonal coefficients of $C(Z)$ to be close to 0, thus decorrelating the different dimensions of the embeddings and preventing them from encoding similar information. The regularization term for textual prompts $r(Z^{\prime})$ can be computed by the same mechanism.

Covariance term increases the diversity of the model's learned features both within individual groups and across the entire ensemble. By implementing this loss across all groups, we enhance diversity within and between groups, fostering varied learning paths and feature exploration. This approach enables groups to develop distinct, yet complementary insights, thereby enriching the model's overall knowledge.
To sum up, the overall objective function $L_{\text{final}}$ can be represented as follows:
\begin{equation*}
L_{\text{final}} = L_{\text{CE-Group1}} + L_{\text{CE-Group2}} + \lambda \left(r(Z)+r(Z^{\prime})\right)
\end{equation*}
where $\lambda$ is the balancing hyper-parameter.

\subsection{Fully Ensemble for Inference}
\noindent{\textbf{Ensemble Inference}} Our inference method is fundamentally based on CLIP's inference approach, which computes the similarity between special tokens $v_0$ and $t_0$ in Eq (\ref{eq:final_output}). To fully leverage the knowledge in both the main prompts and the original CLIP model, we introduce an ensemble inference technique that combines all main prompts in each encoder and utilizes each pair as a classifier, as shown in Figure \ref{fig:overview}:
\begin{equation}
\resizebox{0.9\linewidth}{!}{$
 p(y_k \mid x) = \frac{1}{(2K+1 )^2}\sum_{i=0}^{2K}\sum_{j=0}^{2K}  \frac{\exp (\operatorname{sim}_{i,j}(x, y_k) / \tau)}{\sum_{k'=1}^C \exp (\operatorname{sim}_{i,j}(x, y_{k'}) / \tau)} 
 $}
 \label{eq:postsoftmax}
\end{equation}
where $\operatorname{sim}_{i,j}(x, y_k) = \frac{v_i \cdot t_j}{\left|v_i\right|\left|t_j\right|}$. 

\noindent{\textbf{Special Tokens for Zero-shot Performance}} As shown in Eq \ref{eq:postsoftmax}, our ensemble approach utilizes both groups' prompts and special tokens during inference. Specifically, we employ a special token from CLIP to guide the model in predicting with its pre-trained knowledge. Leveraging these special tokens, we enhance the ensemble mechanism by integrating newly acquired insights into the generalizable knowledge special tokens represent. This configuration is strategically designed to preserve zero-shot capabilities while fostering collaboration between the first and second prompt groups. Such collaborative effort proves particularly beneficial in complex scenarios.


\section{Experiments}
\label{sec:blind}
\subsection{Experiment Setup}
\noindent\textbf{Dataset} For base-to-new generalization and cross-dataset evaluation, we use the 11 image recognition datasets as in~\cite{cocoop}. The datasets include ImageNet~\cite{imagenet} and Caltech101~\cite{caltech} for classification on generic objects; OxfordPets~\cite{pets}, StandfordCars~\cite{stanfordcars}, Flowers102~\cite{flowers}, Food101~\cite{food101} and FGVCAircraft~\cite{aircraft}
 for fine-grained classification; SUN397~\cite{sun397} for scene recognition; UCF101~\cite{ucf101} for action recognition; DTD~\cite{dtd} for texture classification; and EuroSAT~\cite{eurosat} for satellite images. For domain generalization benchmark, we use ImageNet~\cite{imagenet} as a source dataset and ImageNetV2~\cite{imagenetv2}, ImageNet-Sketch~\cite{imagenetsketch}, ImageNet-A~\cite{imagenetA} and ImageNet-R~\cite{imagenetR} as target datasets.  \\

\vspace{-10pt}
\noindent\textbf{Implementation Details} All experiments are based on the vision transformer backbone ViT-B/16 on CLIP~\cite{vit}. Following~\cite{rpo}, we use a total of 24 prompts, with 9 prompts allocated for each of the first/second prompt groups, along with an additional 6 auxiliary prompts$(K=9, K^{\prime}=6)$ for each encoder. 
We train the model with a batch size of 4 and utilize the SGD optimizer, using a learning rate of 0.01. The epochs for each experiment are set as follows: For the base-to-new setting, ImageNet is trained for 15 epochs, while all other datasets are trained for 30 epochs. In extended cross-dataset experiments and domain generalization, all datasets, including ImageNet, are trained for 15 epochs. During training, we use 16 samples per class and set $\lambda$ = 500 to weight the covariance loss. We report accuracy averaged over 3 runs, using the model from the last epoch for evaluation. We conduct overall experiments using a single RTX 3090 GPU. \\
\vspace{-10pt}

\begin{table}[tb]
\centering
\small
\setlength{\tabcolsep}{4pt}
\renewcommand{\arraystretch}{0.9}
\caption{\textbf{Base-to-new generalization.} Trained with a 16-shot setting on base classes and evaluated across both base and novel classes, GPE secures the highest Harmonic mean(H) of base and novel class performances, reflecting its robust generalization capability. The average rank is measured based on the harmonic mean rank for each dataset. The performance of ProGrad is reported from the results in \cite{kgcoop}.}
\label{tab:basetonew}
\resizebox{0.5\textwidth}{!}
{
\begin{tabular}{lc|cccccc|c}
\hline
\multicolumn{2}{l|}{Dataset} &
  CLIP &
  CoOp &
  CoCoOp &
  ProGrad &
  RPO &
  MaPLe &
  \textbf{\begin{tabular}[c]{c}GPE\\(Ours)\end{tabular}} \\ \hline
\rowcolor[HTML]{DEDDDD} 
\multicolumn{2}{c|}{\textbf{Avg. Rank}}       & 6.09           & 6.45           & 4.36           & 4.36     & 3.00  & 2.18      & \textbf{1.55}  \\ \hline
\multirow{2}{*}{\begin{tabular}[c]{@{}l@{}}Average on \\ 11 datasets\end{tabular}} &
  Base &
  69.34 &
  82.69 &
  80.47 &
  82.48 &
  81.13 &
  82.28 &
  \textbf{83.26} \\
  & New  & 74.22  & 63.22 & 71.69 & 70.75& 75.00 & 75.14 & \textbf{75.92} \\
  & H  & 71.70  & 71.66 & 75.83 & 76.16& 77.78 & 78.55 & \textbf{79.24}  \\ \hline
\multirow{3}{*}{ImageNet}     & Base & 72.43          & 76.47          & 75.98     & \textbf{ 77.02 }  & 76.60 & 76.66 & 76.77 \\
 & New  & 68.14 & 67.88  & 70.43  &66.66& \textbf{71.57} & 70.54 & \textbf{71.57 }\\
 & H  & 70.22 & 71.92 & 73.10 &71.46& 74.00 & 73.47 & \textbf{74.08} \\ \hline
\multirow{3}{*}{Caltech101}   & Base & 96.84          & 98.00          & 97.96      &98.02    & 97.97          & 97.74  & \textbf{98.63}\\
& New  & 94.00 & 89.81 & 93.81 & 93.89& \textbf{94.37} & 94.36 & 93.90 \\
& H  & 95.40  & 93.73  & 95.84 & 95.91& 96.03 & 96.02 & \textbf{96.21} \\ \hline
\multirow{3}{*}{OxfordPets}   & Base & 91.17          & 93.67          & \textbf{95.20} & 95.07& 94.63          & 95.43 & 94.60  \\
  & New  & 97.26          & 95.29          & 97.69 & 97.63& 97.50          & 97.76  & \textbf{97.93}         \\
  & H  & 94.12          & 94.47          & 96.43 & 96.33& 96.05          & \textbf{96.58}   &96.24      \\ \hline
\multirow{3}{*}{StanfordCars} & Base & 63.37          & \textbf{78.12} & 70.49      &77.68    & 73.87          & 72.94 &76.47         \\
& New  & 74.89          & 60.40          & 73.59    &68.63      & \textbf{75.53} & 74.00 & 74.50          \\
& H  & 68.65          & 68.13          & 72.01     &72.88     & 74.69          & 73.47 & \textbf{75.47} \\ \hline
\multirow{3}{*}{Flowers102}   & Base & 72.08          & 97.60 & 94.87     &95.54     & 94.13          & 95.92 &\textbf{97.57}          \\
& New  & \textbf{77.80} & 59.67          & 71.75     &71.87     & 76.67          & 72.46 & 76.30        \\
& H  & 74.83          & 74.06          & 81.71     &82.03     & 84.50          & 82.56 & \textbf{85.63} \\ \hline
\multirow{3}{*}{Food101}      & Base & 90.10          & 88.33          & 90.70 &90.37 & 90.33          & \textbf{90.71} &90.60          \\
& New  & 91.22          & 82.26          & 91.29 & 89.59& 90.83          & \textbf{92.05} &91.20          \\
& H  & 90.66          & 85.19          & \textbf{90.99} & 89.98& 90.58          & 91.38 & 90.90          \\ \hline
\multirow{3}{*}{FGVCAircraft} & Base & 27.19          & 40.44 & 33.41    &\textbf{40.54}      & 37.33          & 37.44 & 40.30   \\
& New  & \textbf{36.29} & 22.30          & 23.71      &27.57    & 34.20          & 35.61 & 34.40          \\
& H  & 31.09          & 28.75          & 27.74    &32.82      & 35.70          & 36.50 & \textbf{37.12}\\ \hline
\multirow{3}{*}{SUN397}       & Base & 69.36          & 80.60          & 79.74   &81.26       & 80.60          & 80.82 & \textbf{81.63}\\
& New  & 75.35          & 65.89          & 76.86   &74.17       & 77.80          & 78.70 & \textbf{78.77} \\
& H  & 72.33          & 72.51          & 78.27      &77.55    & 79.18          & 79.75 & \textbf{80.17} \\ \hline
\multirow{3}{*}{DTD}          & Base & 53.24          & 79.44          & 77.01   &77.35       & 76.70          & 80.36 & \textbf{82.07} \\
& New  & 59.90          & 41.18          & 56.00     &52.35     & 62.13          & 59.18 & \textbf{63.97} \\
& H  & 56.37          & 54.24          & 64.85    &62.45      & 68.61          & 68.16 & \textbf{71.89} \\ \hline
\multirow{3}{*}{EuroSAT}      & Base & 56.48          & 92.19 & 87.49   &90.11       & 86.63          & \textbf{94.07} & 92.30          \\
& New  & 64.05          & 54.74          & 60.04    &60.89      & 68.97 & 73.23 & \textbf{76.27}          \\
& H  & 60.03          & 68.69          & 71.21    &72.67      & 76.79 & 82.35 & \textbf{83.52}          \\ \hline
\multirow{3}{*}{UCF101}       & Base & 70.53          & 84.69          & 82.33    &84.33      & 83.67          & 83.00 & \textbf{84.93} \\
& New  & 77.50 & 56.06          & 73.45        &74.94  & 75.43          & \textbf{78.66} & 76.37          \\
& H  & 73.85          & 67.46          & 77.64     & 79.35     & 79.34          & \textbf{80.77} & 80.42 \\ \hline
\end{tabular}}
\vspace{-10pt}
\end{table}


\begin{table*}[tp]
\setlength{\tabcolsep}{3pt}
\renewcommand{\arraystretch}{0.6}
\caption{\textbf{Extended cross-dataset transfer.} The model was trained on all classes of the source dataset and evaluated on the remaining 10 target datasets. The results show the average target performance across 11 source datasets. For a fair comparison, the zero-shot performance of CLIP for each source dataset is also averaged across the other 10 datasets, excluding the source.}
\label{tab:crossdataset}
\centering
\resizebox{0.8\textwidth}{!}{
\setlength{\tabcolsep}{4pt}
\renewcommand{\arraystretch}{1.3}
\begin{tabular}{lcccccccccccc|c}
\hline
\multicolumn{1}{c}{\multirow{6}{*}{{\makecell{Source}}}} & \vspace{-1em} & & & & & & & & & & & & \\ 
& \rotatebox{60}{ImageNet} & \rotatebox{60}{Caltech101} & \rotatebox{60}{OxfordPets} & \rotatebox{60}{StanfordCars} & \rotatebox{60}{Flowers102} & \rotatebox{60}{Food101} & \rotatebox{60}{FGVCAircraft} & \rotatebox{60}{SUN397} & \rotatebox{60}{DTD} & \rotatebox{60}{EuroSAT} & \rotatebox{60}{UCF101} & \rotatebox{60}{\textit{Avg}} & \rotatebox{60}{\textit{Avg. Rank}} \\
\cline{2-14}
\multicolumn{1}{c}{}       & \multicolumn{13}{c}{Target Average except Source}                                                                                          \\ \hline
Zero Shot                                   & 65.10    & 62.48      & 62.86      & 65.24        & 64.64      & 63.16   & 69.29        & 65.51  & 67.32 & 67.00   & 65.10  & 65.25    & -    \\ \hline
CoOp                                      & 63.47    & 49.08      & 44.80      & 46.10        & 39.53      & 43.27   & 48.86        & 53.17  & 50.17 & 40.50   & 41.67  & 47.33    &5.00    \\
CoCoOp                                      & 65.45    & 60.09      & 55.68      & 57.11        & 52.13      & 59.44   & 51.46        & 61.56  & 60.41 & 52.43   & 56.35  & 57.46  &3.82      \\
RPO                                         & 65.21    & 61.89      & 62.13      & \textbf{65.04}        & 58.60      & 62.46   & 65.98        & 63.11  & 59.92 & \textbf{65.92 }  & 60.33  & 62.78    &2.36    \\ 
MaPLe                                         & \textbf{66.20  }  & \textbf{64.63}      & 59.16      & 60.01        & 53.78      & 62.13   & 58.18        & \textbf{64.35}  & \textbf{62.33} & 54.32   & 60.49  & 60.51 &2.18        \\ 
\rowcolor[HTML]{DEDDDD} GPE& 65.79    & 62.57      & 60.61      & 64.52        & \textbf{59.64 }     & \textbf{62.69}   & \textbf{66.94 }       & 64.25  & 62.21 & 64.44   & \textbf{61.27 } & \textbf{63.17}        &\textbf{1.64}\\ \hline
\end{tabular}} 
\end{table*}


\subsection{Base-to-New Generalization}
\noindent\textbf{Setting} We equally split the datasets into base and novel classes, training the model on the base classes in a few-shot setting and evaluating it on both base and novel categories. This benchmark aims to evaluate our approach's within-dataset generalization capabilities. We evaluate GPE against baselines including CoOp~\cite{coop}, CoCoOp~\cite{cocoop}, ProGrad~\cite{prograd}, MaPLe~\cite{maple} and RPO~\cite{rpo}, which served as an inspiration for the masked attention mechanism. 

\noindent\textbf{Result} As shown in Table~\ref{tab:basetonew}, GPE emerges as a standout performer in both base and novel class categories, securing the top harmonic mean compared to baselines. 

GPE significantly outperforms established baselines, particularly in handling novel class scenarios. Many models like CoOp, CoCoOp, and ProGrad struggle to maintain the zero-shot performance of the original CLIP, limiting their generalizability. In contrast, GPE surpasses zero-shot CLIP in novel class evaluations, achieving a 1.7\% improvement in accuracy for novel classes. This enhancement is attributed to our strategic use of masked attention to mitigate internal representation shifts within frozen CLIP, as well as the incorporation of special tokens during inference to preserve zero-shot capabilities.

Additionally, GPE excels in base class performance, surpassing CoOp, which previously held the highest base class accuracy among baseline models. By employing grouping strategies and ensemble methods, GPE improves base class performance by 0.57\% over CoOp. This demonstrates the effectiveness of GPE’s prompt grouping, which not only preserves but also enhances the model's ability to retain and utilize learned knowledge.

\subsection{Extended Cross-Dataset Evaluation}\label{sec:extended_xd}
\noindent\textbf{Setting} We evaluate our model's ability to generalize across different datasets while preserving pretrained knowledge. After training on all classes of one source dataset, we test the model on the remaining 10 target datasets. Instead of previous approach with a single source dataset, such as ImageNet \cite{imagenet}, we extend cross-dataset evaluation by varying the source dataset, each with diverse characteristics and classes. Our extended cross-dataset setting includes fine-grained datasets to assess the model's resilience and its ability to retain general knowledge even after training on specialized data. 
This setup can be extended into a continual learning framework~\cite{cl_zscl, cl_tang2024mind,cl_yu2024boosting}, where the model retains its zero-shot capabilities while preventing the forgetting of knowledge learned from previous datasets and simultaneously acquiring new knowledge from additional datasets. 

\noindent\textbf{Result} In Table~\ref{tab:crossdataset}, we compare GPE’s performance with CLIP zero-shot performance, CoOp, CoCoOp, RPO, and MaPLe. The original CLIP performance on the target dataset serves as the upper bound, indicating that there is a trade-off between preserving zero-shot capability and adapting to new dataset. GPE achieves the highest cross-dataset transfer performance, showing minimal loss in zero-shot capabilities after fine-tuning. Our extended evaluation highlights significant performance drops in other models when fine-tuned on specialized datasets. CoOp and CoCoOp, for example, show substantial declines in performance when fine-tuned on datasets like Flowers102 and FGVCAircraft and evaluated on other datasets. In contrast, GPE, even when fine-tuned on niche datasets, maintains near zero-shot performance, demonstrating its ability to retain knowledge across various training scenarios.

\begin{table}[tb]
\centering
\setlength{\tabcolsep}{4.5pt}
\renewcommand{\arraystretch}{1.0}
\caption{\textbf{Domain generalization setting.} The source model is trained on ImageNet and then evaluated on ImageNet variants.}
\label{tab:domaingeneralization}
\small
\begin{tabular}{lccccc}
\toprule
        & Source                        & \multicolumn{4}{c}{Target}                                                                                                                               \\ \cmidrule(r){2-2} \cmidrule(l){3-6}
        & ImageNet(IN)                      & IN-V2                             & IN-Sketch               & IN-A                             & IN-R                             \\ \bottomrule
CLIP    & 66.73                         & 60.83                                  & 46.15                         & 47.77                                  & 73.96                                  \\
+CoOP   & 71.51                         & 64.20                                  & 47.99                         & 49.71                                  & 75.21                                  \\
+CoCoOp & 71.02                         & 64.07                                  & 48.75                         & 50.63                         & 76.18                                  \\
+RPO    & \cellcolor[HTML]{FFFFFF}\textbf{71.67} & \cellcolor[HTML]{FFFFFF}\textbf{65.13} & \cellcolor[HTML]{FFFFFF}49.27 & \cellcolor[HTML]{FFFFFF}50.13 & \cellcolor[HTML]{FFFFFF}76.57 \\
+MaPLe &70.72 &64.07 &49.15 &\textbf{50.90} &\textbf{76.98 }\\
\rowcolor[HTML]{DEDDDD}  \toprule
GPE    & 71.60  & 64.93   & \textbf{49.30}  &  49.40  & 76.73                                        \\ \bottomrule
\end{tabular}
\vspace{-5pt}
\end{table}


\subsection{Domain Generalization}
\noindent\textbf{Setting} In this evaluation framework, our primary objective is to assess the model's ability to effectively handle out-of-distribution data. The source model is trained on ImageNet~~\cite{imagenet} and then evaluated on four distinct ImageNet variants, which introduce various types of domain shifts. 

\noindent\textbf{Result} As presented in Table~\ref{tab:domaingeneralization}, all methods show improved performance on the out-of-distribution dataset compared to zero-shot CLIP, indicating that the distribution shift is less pronounced than in section \ref{sec:extended_xd} setting. Thus, our strategy, which leverages the knowledge of the original CLIP model, does not show dramatic improvements. However, the results reflect our goal of effectively utilizing existing knowledge while adapting to new datasets, as evidenced by the performance enhancement on the source dataset and comparable performance on the target dataset.

\section{Analysis}
\setlength{\tabcolsep}{2.2pt}
\renewcommand{\arraystretch}{0.95}
\begin{table}[tb]
\centering
\caption{\textbf{Ablation Study on Prompt Diversification Methods.} Averaged results over 11 datasets. The Vendi Score(VS) quantifies the diversity of text prompts embeddings. The $\vartriangle$ indicates the change in VS compared to the default GPE setting.}
\label{prompt diversification}
\centering
\begin{tabular}{l|cc|c|cc}
\toprule
 \small{Method}& \small{Base}  & \small{New}   & \small{H}   & \small{VS} & $\vartriangle$\\\midrule
 
 \rowcolor[HTML]{DEDDDD} \small{GPE}& \textbf{\small{83.26}}& \textbf{\small{75.92}}&\textbf{\small{79.24}} &\textbf{\small{4.593}} & - \\
           \small{GPE w/o grouping\&cov loss} & \small{82.54} & \small{74.46} & \small{78.05} & \small{3.920} & \small{$\vartriangle$0.673}\\
           \small{GPE w/o grouping} & \small{82.65} & \small{75.15} & \small{78.51} &\small{4.007} & $\vartriangle$\small{0.586} \\
 \small{GPE w/o cov loss} & \small{83.24} & \small{75.38} &\small{78.91} &\small{4.447} & \small{\small{$\vartriangle$0.146}} \\

 \bottomrule
\end{tabular}
\end{table}

\subsection{Impact of Prompt Diversification}
In Table~\ref{prompt diversification}, we investigate the prompt diversification effect. To more accurately quantify prompt diversity, we employ the Vendi Score~\cite{vendiscore}, a metric defined as the exponential of the Shannon entropy of the eigenvalues of a similarity matrix. This could reflects the variability across prompts, compared to simpler metrics like L2 distance. We measured the Vendi Score between text prompts embeddings across 11 datasets, using cosine similarity as the similarity function for the Vendi Score, with higher Vendi Scores indicating greater diversity.

To demonstrate the impact of prompt grouping, we compare the total cross-entropy loss with our group-wise cross-entropy loss. This indicates that group-wise learning enhances performance by better utilizing prompt diversification. These findings, detailed in Table~\ref{prompt diversification}, highlight the importance of both grouping and covariance loss in achieving high prompt diversity and reducing information redundancy between prompts. Additionally, we show that employing covariance loss to enhance prompt diversity within and across groups boosts performance. Our results demonstrate that GPE, utilizing both the grouping strategy and covariance loss, yields the highest performance and diversity.

    

\subsection{Effectiveness of Prompt Ensemble Strategies}

\noindent {\bf Analysis on Auxiliary Prompts} To assess the impact of auxiliary prompts, we conduct an ablation study by varying the number of prompt tokens. The results summarized in Table \ref{tab:ablation} show that integrating auxiliary prompts enhances performance compared to using only 18 main prompts without auxiliary prompts ($K=9$). For a fair comparison, we also test 24 prompts without auxiliary prompts ($K=12$). This results in performance deterioration(77.98), highlighting the need for the auxiliary prompts.

Additionally, allowing interactions between the auxiliary prompts and both the first and second groups resulted in a harmonic mean drop to 78.32\%. This demonstrates that our strategic choice to restrict auxiliary prompts to attend only to the second group encourages each group to evolve in diverse and complementary directions.

\noindent\textbf{The Effect of Group-wise Ensemble}
We compare our group-wise ensemble strategy explained in section \ref{sec:group-wise ensemble} with pair-wise training where 18 pairs of main prompts from the text and image encoders generate classifiers, as in RPO. The key difference between pair-wise and group-wise training is that, in pair-wise training, only specific pairs of prompts are used, while in group-wise training, all possible combinations within each group are considered. Pair-wise training results in a performance decline, as shown in Table \ref{tab:ablation}. This is due to fewer classifiers available for ensemble learning compared to our group-wise strategy.

\noindent\textbf{Role of the Special Tokens during Inference}  
We examined the model's performance without the special token, using only 18 main prompts for prediction. Although base class performance showed a slight improvement, new class performance declined, leading to a lower harmonic mean. This highlights the importance of the special token during inference for improving zero-shot generalization.


\noindent {\bf GPE to Single Modality}
Our experiments show that applying GPE solely to the text encoder nearly matches dual-modality performance, while using it only on the image encoder reduces it to 75.32\%, as shown in Table \ref{tab:ablation}. This underscores the critical role of text-based prompting and the importance of balanced prompting across modalities for optimal performance.

\noindent {\bf Number of Groups} We conducted experiments to assess the impact of increasing the number of groups in GPE. Specifically, we created two groups of auxiliary prompts and split the main prompts into three groups: the second group interacts with the first auxiliary group, and the third group interacts with the second auxiliary group. We analyze that expanding the number of groups does not sufficiently diversify the learning trajectories between prompts to improve performance, as shown in Table \ref{tab:ablation}.

\noindent {\bf Prompt Length} As shown in Table \ref{tab:ablation}, we observe a slight performance degradation in both base and novel classes when reducing the total prompt length $K_\text{total}$. Increasing the prompt length resulted in a more significant performance drop, particularly in the novel classes. This degradation is likely due to overfitting to specific domain knowledge as the prompt length increases, reducing the model's ability to generalize effectively for zero-shot learning. Through extensive experiments with various prompt lengths, we determined that setting $K_\text{total}$ to 24 is optimal for balancing performance. 

\begin{table}[tb]
\centering
\setlength{\tabcolsep}{5pt}
\renewcommand{\arraystretch}{0.85}
{\caption{\textbf{Ablation study on GPE's prompt ensemble strategies.} This table compares different configurations of GPE, including the absence of auxiliary prompts ($K^{\prime}=0$) and variations in the total number of prompts ($K_\text{total}$). In the default setting, $K_\text{total}$ is 24.}
\label{tab:ablation}{
\centering
\begin{tabular}{l|cc|c} 
\toprule
Method           & Base  & Novel & H  \\ \midrule
\rowcolor[gray]{0.9}GPE($K=9,K^{\prime}=6$)               & \textbf{83.26} & \textbf{75.92} &\textbf{79.24}\\ 
    GPE ($K=9,K^{\prime}=0$) & 82.68 & 74.19 & 77.99 \\
    GPE ($K=12,K^{\prime}=0$) & 82.60 & 74.23 & 77.98 \\ 
    \midrule
    GPE pair-wise training              & 82.35 & 75.17 & 78.30 \\
    GPE w/o special tokens       & 83.42 & 75.66 & 79.15 \\
        \midrule
Text-GPE          & 82.54 & 75.57 & 78.70 \\
Image-GPE         & 76.40 & 74.47 & 75.32 \\ \midrule
GPE w/ more groups   & 83.44 & 75.53 & 79.08 \\ 
GPE ($K_\text{total}=16$)         & 82.59 & 75.04 & 78.63 \\
GPE ($K_\text{total}=32$)         & 82.54 &73.64  &77.84  \\
    \bottomrule

\end{tabular}
\vspace{-9pt}
}}
\end{table}

\section{Conclusion}
We introduced the Group-wise Prompt Ensemble (GPE) methodology to enhance the adaptability and domain-specific performance of vision-language models like CLIP, preserving their zero-shot learning capabilities. We present a straightforward yet impactful advancement in prompt-based learning techniques by employing a prompt grouping strategy with masked attention and an ensemble method enriched with covariance loss. This innovative approach safeguards CLIP's zero-shot capability while integrating new domain-specific knowledge, addressing a significant gap between adaptability and generalization.

Comprehensive experiments affirm the effectiveness of GPE, showcasing improved performance in base-to-novel class generalization and remarkable robustness in cross-dataset evaluations. Through this work, we have significantly advanced the field of vision-language models, offering a scalable and effective strategy for fine-tuning while maintaining generalization in real-world scenarios.

\paragraph*{Acknowledgements} 
This work was supported by the National Research Foundation of Korea (NRF) grant (No. RS-2023-00222663, RS-2024-00345809), the Institute of Information \& Communications Technology Planning \& Evaluation (IITP) grant (No. RS-2023-00232046, under the Leading Generative AI Human Resources Development, IITP-2025-RS-2024-00397085), both funded by the Korea government (MSIT), and by the Technology Innovation Program (RS-2024-00468747, Development of AI and Lightweight Technology for Embedding Multisensory Intelligence Modules) funded by the Ministry of Trade, Industry \& Energy (MOTIE, Korea).


%

{\small
\bibliographystyle{ieee_fullname}
\bibliography{egbib}

\begin{thebibliography}{10}\itemsep=-1pt

\bibitem{zeroshot_prompt_ensemble}
James~Urquhart Allingham, Jie Ren, Michael~W Dusenberry, Xiuye Gu, Yin Cui, Dustin Tran, Jeremiah~Zhe Liu, and Balaji Lakshminarayanan.
\newblock A simple zero-shot prompt weighting technique to improve prompt ensembling in text-image models.
\newblock In {\em International Conference on Machine Learning}, pages 547--568. PMLR, 2023.

\bibitem{vicreg}
Adrien Bardes, Jean Ponce, and Yann LeCun.
\newblock Vicreg: Variance-invariance-covariance regularization for self-supervised learning.
\newblock {\em arXiv preprint arXiv:2105.04906}, 2021.

\bibitem{food101}
Lukas Bossard, Matthieu Guillaumin, and Luc Van~Gool.
\newblock Food-101--mining discriminative components with random forests.
\newblock In {\em Computer Vision--ECCV 2014: 13th European Conference, Zurich, Switzerland, September 6-12, 2014, Proceedings, Part VI 13}, pages 446--461. Springer, 2014.

\bibitem{gpt3}
Tom Brown, Benjamin Mann, Nick Ryder, Melanie Subbiah, Jared~D Kaplan, Prafulla Dhariwal, Arvind Neelakantan, Pranav Shyam, Girish Sastry, Amanda Askell, et~al.
\newblock Language models are few-shot learners.
\newblock {\em Advances in neural information processing systems}, 33:1877--1901, 2020.

\bibitem{apollo}
Sanjoy Chowdhury, Sayan Nag, and Dinesh Manocha.
\newblock Apollo: Unified adapter and prompt learning for vision language models.
\newblock In {\em The 2023 Conference on Empirical Methods in Natural Language Processing}, 2023.

\bibitem{dtd}
Mircea Cimpoi, Subhransu Maji, Iasonas Kokkinos, Sammy Mohamed, and Andrea Vedaldi.
\newblock Describing textures in the wild.
\newblock In {\em Proceedings of the IEEE conference on computer vision and pattern recognition}, pages 3606--3613, 2014.

\bibitem{imagenet}
Jia Deng, Wei Dong, Richard Socher, Li-Jia Li, Kai Li, and Li Fei-Fei.
\newblock Imagenet: A large-scale hierarchical image database.
\newblock In {\em 2009 IEEE conference on computer vision and pattern recognition}, pages 248--255. Ieee, 2009.

\bibitem{bert}
Jacob Devlin, Ming-Wei Chang, Kenton Lee, and Kristina Toutanova.
\newblock Bert: Pre-training of deep bidirectional transformers for language understanding.
\newblock {\em arXiv preprint arXiv:1810.04805}, 2018.

\bibitem{dietterich2000ensemble}
Thomas~G Dietterich.
\newblock Ensemble methods in machine learning.
\newblock In {\em International workshop on multiple classifier systems}, pages 1--15. Springer, 2000.

\bibitem{vit}
Alexey Dosovitskiy, Lucas Beyer, Alexander Kolesnikov, Dirk Weissenborn, Xiaohua Zhai, Thomas Unterthiner, Mostafa Dehghani, Matthias Minderer, Georg Heigold, Sylvain Gelly, Jakob Uszkoreit, and Neil Houlsby.
\newblock An image is worth 16x16 words: Transformers for image recognition at scale, 2021.

\bibitem{caltech}
Li Fei-Fei, Rob Fergus, and Pietro Perona.
\newblock Learning generative visual models from few training examples: An incremental bayesian approach tested on 101 object categories.
\newblock In {\em 2004 conference on computer vision and pattern recognition workshop}, pages 178--178. IEEE, 2004.

\bibitem{deep_ensemble}
Stanislav Fort, Huiyi Hu, and Balaji Lakshminarayanan.
\newblock Deep ensembles: A loss landscape perspective.
\newblock {\em arXiv preprint arXiv:1912.02757}, 2019.

\bibitem{vendiscore}
Dan Friedman and Adji~Bousso Dieng.
\newblock The vendi score: A diversity evaluation metric for machine learning, 2023.

\bibitem{eurosat}
Patrick Helber, Benjamin Bischke, Andreas Dengel, and Damian Borth.
\newblock Eurosat: A novel dataset and deep learning benchmark for land use and land cover classification.
\newblock {\em IEEE Journal of Selected Topics in Applied Earth Observations and Remote Sensing}, 12(7):2217--2226, 2019.

\bibitem{imagenetR}
Dan Hendrycks, Steven Basart, Norman Mu, Saurav Kadavath, Frank Wang, Evan Dorundo, Rahul Desai, Tyler Zhu, Samyak Parajuli, Mike Guo, et~al.
\newblock The many faces of robustness: A critical analysis of out-of-distribution generalization.
\newblock In {\em Proceedings of the IEEE/CVF International Conference on Computer Vision}, pages 8340--8349, 2021.

\bibitem{imagenetA}
Dan Hendrycks, Kevin Zhao, Steven Basart, Jacob Steinhardt, and Dawn Song.
\newblock Natural adversarial examples.
\newblock In {\em Proceedings of the IEEE/CVF Conference on Computer Vision and Pattern Recognition}, pages 15262--15271, 2021.

\bibitem{align}
Chao Jia, Yinfei Yang, Ye Xia, Yi-Ting Chen, Zarana Parekh, Hieu Pham, Quoc Le, Yun-Hsuan Sung, Zhen Li, and Tom Duerig.
\newblock Scaling up visual and vision-language representation learning with noisy text supervision.
\newblock In {\em International conference on machine learning}, pages 4904--4916. PMLR, 2021.

\bibitem{maple}
Muhammad~Uzair Khattak, Hanoona Rasheed, Muhammad Maaz, Salman Khan, and Fahad~Shahbaz Khan.
\newblock Maple: Multi-modal prompt learning.
\newblock In {\em Proceedings of the IEEE/CVF Conference on Computer Vision and Pattern Recognition}, pages 19113--19122, 2023.

\bibitem{promptsrc}
Muhammad~Uzair Khattak, Syed~Talal Wasim, Muzammal Naseer, Salman Khan, Ming-Hsuan Yang, and Fahad~Shahbaz Khan.
\newblock Self-regulating prompts: Foundational model adaptation without forgetting.
\newblock In {\em Proceedings of the IEEE/CVF International Conference on Computer Vision}, pages 15190--15200, 2023.

\bibitem{stanfordcars}
Jonathan Krause, Michael Stark, Jia Deng, and Li Fei-Fei.
\newblock 3d object representations for fine-grained categorization.
\newblock In {\em Proceedings of the IEEE international conference on computer vision workshops}, pages 554--561, 2013.

\bibitem{rpo}
Dongjun Lee, Seokwon Song, Jihee Suh, Joonmyeong Choi, Sanghyeok Lee, and Hyunwoo~J Kim.
\newblock Read-only prompt optimization for vision-language few-shot learning.
\newblock In {\em Proceedings of the IEEE/CVF International Conference on Computer Vision}, pages 1401--1411, 2023.

\bibitem{proda}
Yuning Lu, Jianzhuang Liu, Yonggang Zhang, Yajing Liu, and Xinmei Tian.
\newblock Prompt distribution learning.
\newblock In {\em Proceedings of the IEEE/CVF Conference on Computer Vision and Pattern Recognition}, pages 5206--5215, 2022.

\bibitem{aircraft}
Subhransu Maji, Esa Rahtu, Juho Kannala, Matthew Blaschko, and Andrea Vedaldi.
\newblock Fine-grained visual classification of aircraft.
\newblock {\em arXiv preprint arXiv:1306.5151}, 2013.

\bibitem{flowers}
Maria-Elena Nilsback and Andrew Zisserman.
\newblock Automated flower classification over a large number of classes.
\newblock In {\em 2008 Sixth Indian conference on computer vision, graphics \& image processing}, pages 722--729. IEEE, 2008.

\bibitem{pets}
Omkar~M Parkhi, Andrea Vedaldi, Andrew Zisserman, and CV Jawahar.
\newblock Cats and dogs.
\newblock In {\em 2012 IEEE Conference on Computer Vision and Pattern Recognition}, pages 3498--3505. IEEE, 2012.

\bibitem{clip}
Alec Radford, Jong~Wook Kim, Chris Hallacy, Aditya Ramesh, Gabriel Goh, Sandhini Agarwal, Girish Sastry, Amanda Askell, Pamela Mishkin, Jack Clark, et~al.
\newblock Learning transferable visual models from natural language supervision.
\newblock In {\em International conference on machine learning}, pages 8748--8763. PMLR, 2021.

\bibitem{gpt}
Alec Radford, Jeffrey Wu, Rewon Child, David Luan, Dario Amodei, Ilya Sutskever, et~al.
\newblock Language models are unsupervised multitask learners.
\newblock {\em OpenAI blog}, 1(8):9, 2019.

\bibitem{imagenetv2}
Benjamin Recht, Rebecca Roelofs, Ludwig Schmidt, and Vaishaal Shankar.
\newblock Do imagenet classifiers generalize to imagenet?
\newblock In {\em International conference on machine learning}, pages 5389--5400. PMLR, 2019.

\bibitem{tpt}
Manli Shu, Weili Nie, De-An Huang, Zhiding Yu, Tom Goldstein, Anima Anandkumar, and Chaowei Xiao.
\newblock Test-time prompt tuning for zero-shot generalization in vision-language models.
\newblock {\em Advances in Neural Information Processing Systems}, 35:14274--14289, 2022.

\bibitem{ucf101}
Khurram Soomro, Amir~Roshan Zamir, and Mubarak Shah.
\newblock Ucf101: A dataset of 101 human actions classes from videos in the wild.
\newblock {\em arXiv preprint arXiv:1212.0402}, 2012.

\bibitem{cl_tang2024mind}
Longxiang Tang, Zhuotao Tian, Kai Li, Chunming He, Hantao Zhou, Hengshuang Zhao, Xiu Li, and Jiaya Jia.
\newblock Mind the interference: Retaining pre-trained knowledge in parameter efficient continual learning of vision-language models.
\newblock {\em arXiv preprint arXiv:2407.05342}, 2024.

\bibitem{transformer}
Ashish Vaswani, Noam Shazeer, Niki Parmar, Jakob Uszkoreit, Llion Jones, Aidan~N Gomez, {\L}ukasz Kaiser, and Illia Polosukhin.
\newblock Attention is all you need.
\newblock {\em Advances in neural information processing systems}, 30, 2017.

\bibitem{imagenetsketch}
Haohan Wang, Songwei Ge, Zachary Lipton, and Eric~P Xing.
\newblock Learning robust global representations by penalizing local predictive power.
\newblock volume~32, 2019.

\bibitem{wortsman2022robust}
Mitchell Wortsman, Gabriel Ilharco, Jong~Wook Kim, Mike Li, Simon Kornblith, Rebecca Roelofs, Raphael~Gontijo Lopes, Hannaneh Hajishirzi, Ali Farhadi, Hongseok Namkoong, et~al.
\newblock Robust fine-tuning of zero-shot models.
\newblock In {\em Proceedings of the IEEE/CVF Conference on Computer Vision and Pattern Recognition}, pages 7959--7971, 2022.

\bibitem{sun397}
Jianxiong Xiao, James Hays, Krista~A Ehinger, Aude Oliva, and Antonio Torralba.
\newblock Sun database: Large-scale scene recognition from abbey to zoo.
\newblock In {\em 2010 IEEE computer society conference on computer vision and pattern recognition}, pages 3485--3492. IEEE, 2010.

\bibitem{kgcoop}
Hantao Yao, Rui Zhang, and Changsheng Xu.
\newblock Visual-language prompt tuning with knowledge-guided context optimization.
\newblock In {\em Proceedings of the IEEE/CVF Conference on Computer Vision and Pattern Recognition (CVPR)}, pages 6757--6767, June 2023.

\bibitem{yao2021filip}
Lewei Yao, Runhui Huang, Lu Hou, Guansong Lu, Minzhe Niu, Hang Xu, Xiaodan Liang, Zhenguo Li, Xin Jiang, and Chunjing Xu.
\newblock Filip: Fine-grained interactive language-image pre-training.
\newblock {\em arXiv preprint arXiv:2111.07783}, 2021.

\bibitem{cl_yu2024boosting}
Jiazuo Yu, Yunzhi Zhuge, Lu Zhang, Ping Hu, Dong Wang, Huchuan Lu, and You He.
\newblock Boosting continual learning of vision-language models via mixture-of-experts adapters.
\newblock In {\em Proceedings of the IEEE/CVF Conference on Computer Vision and Pattern Recognition}, pages 23219--23230, 2024.

\bibitem{yuan2021florence}
Lu Yuan, Dongdong Chen, Yi-Ling Chen, Noel Codella, Xiyang Dai, Jianfeng Gao, Houdong Hu, Xuedong Huang, Boxin Li, Chunyuan Li, et~al.
\newblock Florence: A new foundation model for computer vision.
\newblock {\em arXiv preprint arXiv:2111.11432}, 2021.

\bibitem{barlow}
Jure Zbontar, Li Jing, Ishan Misra, Yann LeCun, and St{\'e}phane Deny.
\newblock Barlow twins: Self-supervised learning via redundancy reduction.
\newblock In {\em International Conference on Machine Learning}, pages 12310--12320. PMLR, 2021.

\bibitem{zhai2022lit}
Xiaohua Zhai, Xiao Wang, Basil Mustafa, Andreas Steiner, Daniel Keysers, Alexander Kolesnikov, and Lucas Beyer.
\newblock Lit: Zero-shot transfer with locked-image text tuning.
\newblock In {\em Proceedings of the IEEE/CVF Conference on Computer Vision and Pattern Recognition}, pages 18123--18133, 2022.

\bibitem{cl_zscl}
Zangwei Zheng, Mingyuan Ma, Kai Wang, Ziheng Qin, Xiangyu Yue, and Yang You.
\newblock Preventing zero-shot transfer degradation in continual learning of vision-language models.
\newblock In {\em Proceedings of the IEEE/CVF International Conference on Computer Vision}, pages 19125--19136, 2023.

\bibitem{cocoop}
Kaiyang Zhou, Jingkang Yang, Chen~Change Loy, and Ziwei Liu.
\newblock Conditional prompt learning for vision-language models.
\newblock In {\em Proceedings of the IEEE/CVF Conference on Computer Vision and Pattern Recognition}, pages 16816--16825, 2022.

\bibitem{coop}
Kaiyang Zhou, Jingkang Yang, Chen~Change Loy, and Ziwei Liu.
\newblock Learning to prompt for vision-language models.
\newblock {\em International Journal of Computer Vision}, 130(9):2337--2348, 2022.

\bibitem{prograd}
Beier Zhu, Yulei Niu, Yucheng Han, Yue Wu, and Hanwang Zhang.
\newblock Prompt-aligned gradient for prompt tuning.
\newblock In {\em Proceedings of the IEEE/CVF International Conference on Computer Vision}, pages 15659--15669, 2023.

\end{thebibliography}
}
\clearpage
\appendix


\end{document}